# Kinodynamic Motion Planning:
# A Novel Type Of Nonlinear, Passive Damping Forces (NADFs)

**Ahmad A. Masoud**
Electrical Engineering Department, KFUPM, P.O. Box 287, Dhaharan 31261, Saudi Arabia, masoud@kfupm.edu.sa

**Abstract-** This paper extends the capabilities of the harmonic potential field approach to planning to cover both the kinematic and dynamic aspects of a robot's motion. The suggested approach converts the gradient guidance field from a harmonic potential to a control signal by augmenting it with a novel type of dampening forces suggested in this paper called: nonlinear, anisotropic, dampening forces (NADFs). The combination of the two provides a signal that can both guide a robot and effectively manage its dynamics. The kinodynamic planning signal inherits, fully, the guidance capabilities of the harmonic gradient field. It can also be easily configured to efficiently suppress the inertia-induced transients in the robot's trajectory without compromising the speed of operation. The approach works with dissipative systems as well as systems acted on by external forces without needing full knowledge of the system's dynamics. Theoretical developments and simulation results are provided in the paper.

## I. Introduction

The harmonic potential field approach to planning is emerging as a powerful paradigm for the guidance of autonomous agents. Since it was suggested in the mid-late eighties [1,2] the approach is continuously developing to meet the stringent requirements operation in a real-life environment imposed on an agent. Up-till-now, the approach has amassed many attractive properties crucial for enhancing goal reachability. The approach is provably-correct driving the agent to a successful conclusion if the task is manageable and providing an indication if the task is intractable. It can be used to guide the motion of an arbitrarily shaped agent in an unknown environment regardless of its geometry or even topology relying only on the sensory data acquired online by the agent's finite range sensors. The method can also impose a variety of constraints on the agent's trajectory such as regional avoidance and directional constraints [3-8]. Harmonic functions are also Morse functions and a general form of the navigation functions suggested in [13], see appendix.

A planner may be defined as an intelligent, purposive, context-sensitive controller that can instruct an agent on how to deploy its motion actuators (i.e generate a control signal) so that a target state may be reached in a constrained manner. Traditionally, a planning task is distributed on two stages: a high level control (HLC) stage and a low level control stage (LLC), figure-1. The HLC stage receives data about the environment, the target of the agent, and constraints on its behavior. It then simultaneously processes these data to generate a reference plan or trajectory marking the desired behavior of the robot. This trajectory, if actualized, leads to the agent reaching its target in the specified manner. The reference trajectory is then fed to an LLC in order to convert it into a sequence of action instructions to be executed by the agent's actuators of motion. Unfortunately, the HLC-LLC paradigm for planning suffers from serious problems that adversely impact its performance in a realistic setting. An alternative may be achieved by fusing the HLC and LLC modules into one called the navigation control (NC). An NC attempts to directly convert the environmental data, goal of the robot, and constraints on its behavior into a control signal (figure-1). Khatib potential field (PF) approach may be considered as one of the first methods to cast planning in an NC framework [9]. The PF approach enjoys several attractive features; most significant is the high speed by which a robot can respond to the contents of its environment.

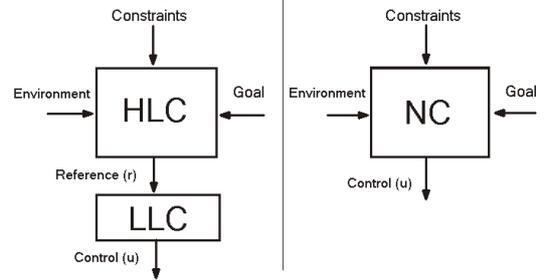

Figure-1: The HLC-LLC and NC control structures.

The attractor-repeller setting Khatib used to generate the potential field has some problems. The most serious one has to do with convergence where it was observed that a robot guided by such a method may stop somewhere in the workspace before reaching its target; the problem was termed the local minima problem. Many methods later appeared to generate potential fields that do not suffer from this problem [10-12]. Koditschek diffeomorphism approach [13] was among the first methods suggested to remedy this shortcoming in the PF approach. To convert the gradient guidance field from the potential surface ($-\nabla V$) into a control signal (u), a viscous dampening force that is linearly proportional to speed is added:

$$u = -B \cdot \dot{x} - \nabla V(x) \qquad (1)$$

This combination will only work provided that the initial speed of the robot is lower than an upper bound $S(x)$:

$$\omega(x) \leq S(x) \qquad x \in \Omega \qquad (2)$$

where $\omega(x)$ is the initial speed of the robot at a location x, and $\Omega$ is the workspace of the robot [14]. Practical application of the above faced two difficulties: first, no method was provided to compute the upper bound S. Even if a method is devised for doing so, there is no guarantee that in a practical situation the initial speed of a robot can be made to lie below the admissible upper bound. The second difficulty has to do with the fact that the satisfaction of the upper speed constraint guarantees only that obstacle avoidance constraints will be upheld and convergence to the target will be achieved. In potential field methods, transients can be a serious concern that could make it impractical to use these techniques for controlling a robot. Also, the approach seems to only deal with dissipative systems where no mention of how the method may be applied when external forces such as gravity are present.

In its current form the HPF approach can only operate in an HLC mode providing only a guidance signal from the gradient of the potential. This signal has to be converted into a control signal by an LLC. Guldner and Utkin suggested an interesting approach based on a sliding mode control for converting the gradient field from an HPF into a control signal [21]. The approach is robust, has good convergence properties, does not require full knowledge of system dynamics and can make, with little transients, the dynamic trajectory of the robot follow the kinematic trajectory marked by the gradient field. The main drawback fo the approach seems to be the high shattering the control signal experiences.

In this paper a method is suggested to utilize the HPF approach in an NC mode. This is accomplished by augmenting the gradient guidance field from an HPF with a new type of dampening force called: nonlinear anisotropic dampening forces (NADFs). It is shown that an NADF-based control can efficiently suppress inertia-induced artifacts in the dynamical trajectory of the system making it closely follow the kinematic trajectory while maintaining an agile system response. The approach does not require the system dynamics to be fully known. A loose upper bound is sufficient for constructing a well-behaved control signal that can deal with dissipative systems as well as systems being influenced by external forces (e.g. gravity).

This paper is organized as follows: section II provides a brief background of the potential field approach. The NADF approach is presented in section III. Sections IV and V discuss the application of the approach to dissipative systems and systems experiencing external forces respectively. Simulation results are in section VI, and conclusions are placed in section VII.

## II. Background

The HPF approach appeared shortly after the work of Khatib. Although the approach was brought to the forefront of motion planning independently and simultaneously by different researchers [16-20], the first work to be published on the subject was that by Sato in 1986 [1]. The HPF approach eliminates the local minima problem encountered in [9] by forcing the differential properties of the potential field to satisfy the Laplace equation inside the workspace of the robot ($\Omega$) while constraining the properties of the potential at the boundary of $\Omega$ ($\Gamma = \partial\Omega$). The boundary set $\Gamma$ includes both the boundaries of the forbidden zones (O) and the target point ($x_T$). A basic setting of the HPF approach is:

$$\nabla^2 V(x) \equiv 0 \qquad x \in \Omega$$
subject to: $\quad V = 0|_{X = X_T} \ \& \ V = 1|_{X \in \Gamma}$ . (3)

The trajectory to the target (x(t)) is generated using the HPF-based, gradient dynamical system:

$$\dot{x} = -\nabla V(x) \qquad x(0) = x_0 \in \Omega \qquad (4)$$

The trajectory is guaranteed to:

1- $\lim_{t \to \infty} x(t) \to x_T$     2- $x(t) \in \Omega \qquad \forall t$     (5)

whereby a proof of (5) may be found in [3]. Figure-2 shows the negative gradient field of a harmonic potential for the simple environment of a room with two dividers. Figure-3 shows the trajectory, x(t), generated using the gradient dynamical system in (4). It ought to be mentioned that the HPF approach is only a special case of a broader class of planners called PDE-ODE motion planners [5] where the field is generated using the boundary value problem (BVP): solve:

$$L(V(x)) \equiv 0 \qquad x \in \Omega$$
subject to: $\quad \chi(V(x)) = 0 \qquad x \in \Gamma.$ (6)

The trajectory is generated using the nonlinear system:

$$\dot{x} = F(V(x)) \qquad x(0) = x_0 \in \Omega \qquad (7)$$

where L is scalar partial differential operator, $\chi$ is a governing relation restricting the potential or some of its properties at the boundary to a certain value, $F$ is a nonlinear vector function mapping $R \to R^N$, N is the dimension of x, PDE stands for partial differential equation, and ODE stands for ordinary differential equation. Planners assuming a PDE-ODE setting other than that of the one in (3) may be found in [3,7,8].

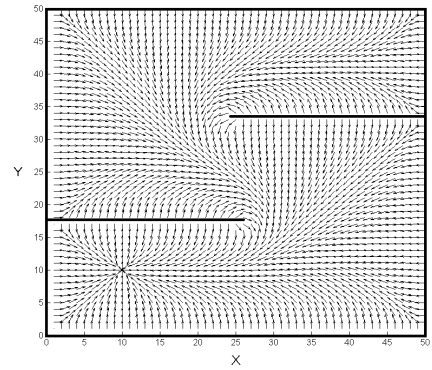

Figure-2: Guidance field of an HPF.

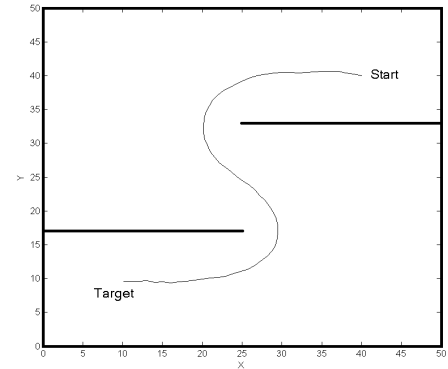

Figure-3: Trajectory generated by the field in figure-2.

The trajectory, x(t), generated by the dynamical system in (4) is only a reference trajectory that should be fed to an LLC in order to generate the control signal, u, the robot is using. One way of converting the guidance signal into a control signal is to augment the gradient field with a component that is proportional to speed. This seemingly straight forward solution is problematic. In figure-4, the negative gradient of the potential in figure-2 is used to navigate a 1kg point mass. The dynamic equation of the system is:

$$\begin{bmatrix} \ddot{x} \\ \ddot{y} \end{bmatrix} = -\mathbf{B} \cdot \begin{bmatrix} \dot{x} \\ \dot{y} \end{bmatrix} - \begin{bmatrix} \partial V(x,y)/\partial x \\ \partial V(x,y)/\partial y \end{bmatrix} \qquad (8)$$

where B=0.1. Despite the fact that the initial speed of the robot is zero, the trajectory violated the avoidance condition and collided with the walls of the room.

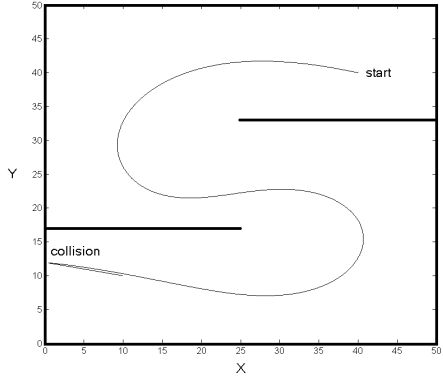

Figure-4: trajectory of a point mass controlled by the field in figure-2.

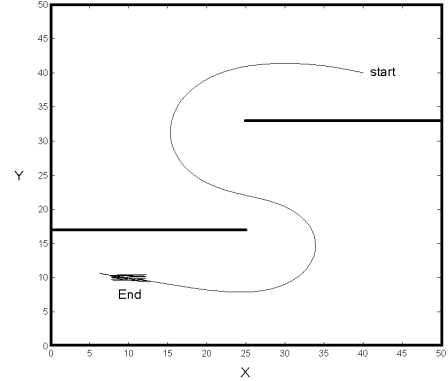

Figure-5: Trajectory, point mass, linear dampening increased.

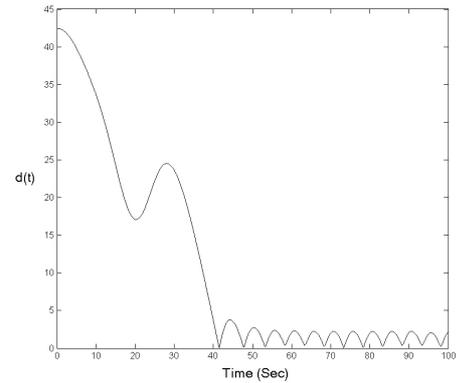

Figure-6: Distance to target versus time.

### III. The NADF Approach

An intuitive solution to the problem of converting a gradient guidance field into a navigation control signal is to increase the coefficient of the linear velocity term to a sufficiently high level. The linear velocity component acts as a dampener of motion that may be used to place the inertial force under control by marginalizing its disruptive influence on the trajectory of the robot that the gradient field is attempting to generate. The following example demonstrates that this solution is impractical. In order to generate a control signal that would satisfy the avoidance constraints (5), the coefficient of dampening of the system is increased to B=0.15. Figure-5 shows the resulting trajectory and figure-6 shows the distance to the target as a function of time. Although the trajectory did converge to the target point ($x_T$) and did not violate the regional avoidance constraints, unacceptable transients along with significant deviations from the path marked by the gradient field (figure-2) are present. In a second attempt to generate a well-behaved control signal, the dampening coefficient is significantly increased to B=.7. Although a well-behaved trajectory was obtained (figure-7), significant slowdown of motion did occur (figure-8).

The method for converting the gradient field from a harmonic potential into a navigation control signal by simple augmentation with a linear velocity dampening term is incorrect. This approach ignores the dual role the gradient field plays as a control and guidance provider. The field guides a robot to the target using vectors that point out the directions along which the robot has to move if the target is to be reached and the obstacles are to be avoided. At the same time, these vectors are forces that act on the mass of the robot in order to actuate motion. Obviously the inertia of the robot will have a disruptive influence on motion. The linear dampening term manages the inertial forces in an attempt to make the motion yield to the guidance provided by the gradient field.

A dampening component that is proportional to velocity exercises omni-directional attenuation of motion regardless of the direction along which it is heading. This means that the useful component of motion marked by the direction along which the goal component of the gradient of the artificial potential is pointing is treated in the same manner as the unwanted inertia-induced, noise component of the trajectory. These two components should not be treated equally. Attenuation should be restricted to the inertia-caused disruptive component of motion, while the component in conformity with the guidance of the artificial potential should be left unaffected (figure-9).

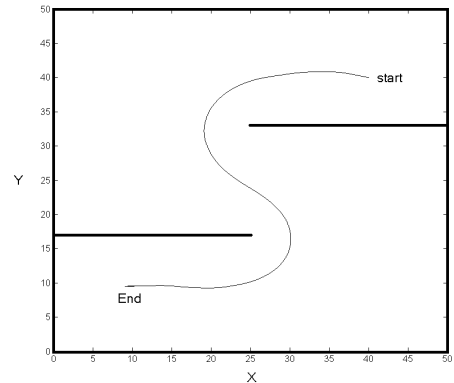

Figure-7: Trajectory, high linear dampening.

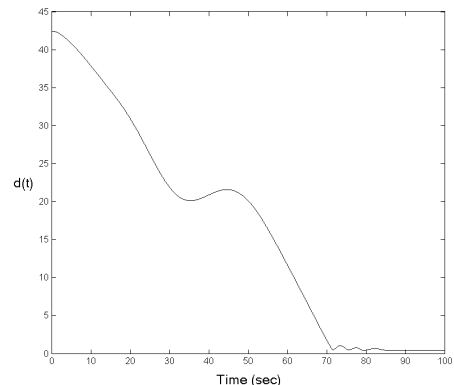

Figure-8: Distance to target versus time.

To better manage the effect of the inertial forces, a more carefully constructed dampening component that treats the gradient of the artificial potential both as an actuator of dynamics and as a guiding signal is needed. A dampening force (M) that behaves in the above manner is:

$$M(x,\dot{x}) = [(n^t \dot{x})n + (\frac{\nabla V(x)^T}{|\nabla V(x)^T|} \cdot \dot{x} \cdot \Phi(\nabla V(x)^T \dot{x}))\frac{\nabla V(x)}{|\nabla V(x)|}] \quad (9)$$

where **n** is a unit vector orthogonal to $\nabla V$. This force is given the name: nonlinear, anisotropic, dampening force (NADF). For the two dimensional case, an NADF has the form:

$$\mathbf{M} = \frac{1}{g_x^2 + g_y^2}\left[(g_x \dot{y} - g_y \dot{x}) \cdot \begin{bmatrix} -g_y \\ g_x \end{bmatrix} + (g_x \dot{x} + g_y \dot{y}) \cdot \Phi(-g_x \dot{x} - g_y \dot{y})\begin{bmatrix} g_x \\ g_y \end{bmatrix}\right] \quad (10)$$

where $\nabla \mathbf{V}(\mathbf{x,y}) = [g_x \quad g_y]^T$.

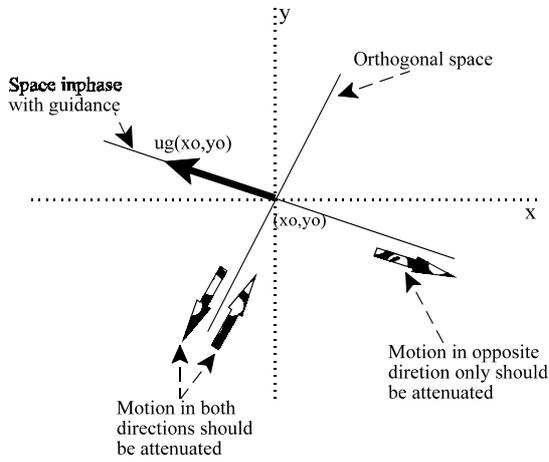

a- action of the dampening force

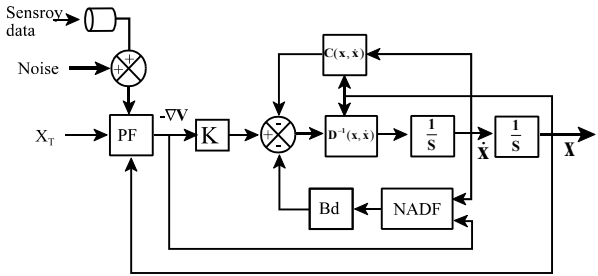

b- block diagram

Figure-9: nonlinear, anisotropic, dampening force (NADF).

## IV- Dissipative Systems

In this section two propositions are stated and proven. The first proposition shows that a gradient field of a harmonic potential generated by the BVP in (3) combined with NADF can guarantee global, asymptotic convergence of a fully actuated second order dissipative dynamical system. The second proposition shows that the dynamic trajectory of the system can be made arbitrarily close to the kinematic trajectory generated by the system in (4); hence, preserving the spatial constraints.

Proposition-1: Let V(x) be a harmonic potential generated using the BVP in (3). The trajectory of the dynamical system:

$$D(x)\ddot{x} + C(x,\dot{x})\dot{x} + B_d \cdot M(x,\dot{x}) + K \cdot \nabla V(x) = 0 \quad (11)$$

will globally, asymptotically converge to:

$$\lim_{t\to\infty}\begin{bmatrix} x \\ \dot{x} \end{bmatrix} \to \begin{bmatrix} x_T \\ 0 \end{bmatrix} \quad (12)$$

for any positive constants $B_d$ and K, where $x \in R^N$, $V(x):R^N \to R$, D(x) is an N×N positive definite inertia matrix, $C(x,\dot{x})\dot{x}$ contains the centripetal, Coriolis, and gyroscopic forces. Proof of the above proposition is carried out using the LaSalle invariance principle [23].

Proof: Let $\Xi$ be the Liapunov function candidate:

$$\Xi(x,\dot{x}) = K \cdot V(x) + \frac{1}{2}\dot{x}^T D(x)\dot{x} \quad (13)$$

Note that since V(x) is harmonic, it must assume its maxima on $\Gamma$ and minima on $x_T$. In other words, V(x) can only be zero at $x_T$; otherwise, its value is greater than zero:

$$\Xi(x,\dot{x}) = \begin{bmatrix} 0 & \text{iff} \quad x=0, \dot{x}=0 \\ \text{positive} & \text{otherwise} \end{bmatrix}. \quad (14)$$

The time derivative of the above function is:

$$\dot{\Xi}(x,\dot{x}) = K \cdot \nabla V(x)^T \dot{x} + \frac{1}{2}\dot{x}^T \dot{D}(x)\dot{x} + \dot{x}^T D(x)\ddot{x}. \quad (15)$$

Substituting:

$$\ddot{x} = D^{-1}(x)[-C(x,\dot{x})\dot{x} - Bd \cdot M(x,\dot{x}) - K \cdot \nabla V(x)] \quad (16)$$

along with (9) in the above equation yields:

$$\begin{aligned}\dot{\Xi} =\ & K \cdot \nabla V(x)^T \dot{x} + \frac{1}{2}\dot{x}^T \dot{D}(x)\dot{x} \\ & - K \cdot \nabla V(x)^T \dot{x} - \dot{x}^T C(x,\dot{x})\dot{x} \\ & - B_d \cdot \dot{x}^T (n^T \dot{x})n \\ & - B_d \cdot \dot{x}^T (\frac{\nabla V(x)^T}{|\nabla V(x)|} \cdot \dot{x} \cdot \Phi(-\nabla V(x)^T \dot{x})\frac{\nabla V(x)}{|\nabla V(x)|}\end{aligned} \quad (17)$$

Using the passivity property:

$$\dot{x}^T(\dot{D}(x) - 2 \cdot C(x,\dot{x}))\dot{x} = 0 \quad (18)$$

and rearranging the terms we get:

$$\begin{aligned}\dot{\Xi} =\ & -B_d \cdot (n^T \dot{x})^T (n^T \dot{x}) \\ & -B_d \cdot \frac{(\nabla V(x)^T \cdot \dot{x})^T}{|\nabla V(x)|} \cdot \frac{(\nabla V(x)^T \cdot \dot{x})}{|\nabla V(x)|} \cdot \Phi(\nabla V(x)^T \dot{x})\end{aligned} \quad (19)$$

as can be seen

$$\dot{\Xi} \le 0 \quad \forall\ x,\dot{x}, \quad (20)$$

where $\dot{\Xi} = 0 \quad \text{for}\ x,\dot{x} = 0$

according to LaSalle principle any bounded solution of (11) will converge to the minimum invariant set:

$$E \subset \{\dot{x}=0, x\}. \quad (21)$$

Determining E requires studying the critical points of V(x) where $\nabla V(x)=0$. According to the maximum principle, $x_T$ is the only minimum (stable equilibrium point) V(x) can have. Besides $x_T$, V(x) has other critical points $\{x_i\}$ at which $\nabla V=0$; however, the hessian at these points is non-singular, i.e. V(x) is Morse [24]. A proof of this result may be found in the appendix. From the above we conclude that E contains only one point which is the point $x = x_T, \dot{x} = 0$ to which motion will converge.

Proposition-2: Let ρ be the trajectory constructed as the spatial projection of the solution, x(t), of the first order differential system in (4). Also Let $ρ_d$ be the trajectory constructed as the spatial projection of the solution, x(t), of the second order system in (11), figure-10. Then there exist a Bd that can make the maximum deviation between ρ and $ρ_d$ ($δ_m$) arbitrarily small.

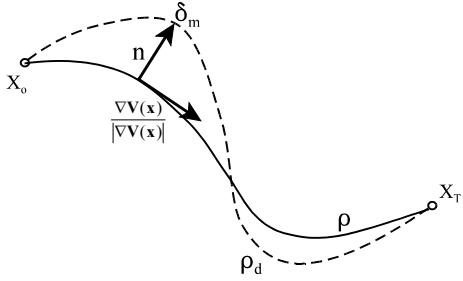

Figure-10: The kinematic and dynamic trajectories.

Proof: The gradient field from an HPF does not only work as a guide of motion to the target; it also may be used to cover Ω with a complete set of boundary-fitted basis coordinates (figure-11).

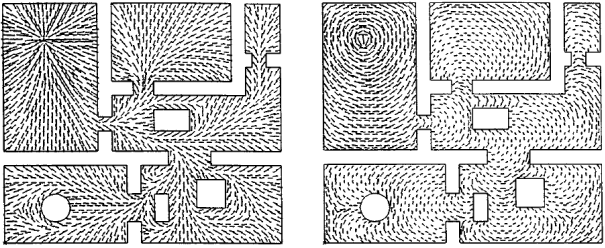

Figure-11: Boundary-fitted coordinate system.

The radial basis of the system ($∇V/|∇V|$) marks the useful component of motion. The basis orthogonal to this component span the disruptive component of motion (δ) which NADF is required to attenuate (figure-12).

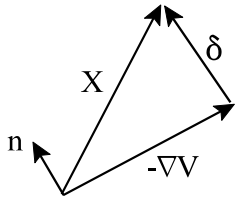

Figure-12: The disruptive component of motion.

The dynamic equation describing the disruptive component is:

$$n^T D(x)\ddot{x} + n^T C(x,\dot{x})\dot{x} + Bd \cdot n^T M(x,\dot{x}) + K \cdot n^T ∇V(x) = 0 \quad (25)$$

Examining the above equation term by term yields:
1- $n^T ∇V = 0$, (26)
2- $n^T[(n^t \dot{x})n + (\frac{∇V(x)^T}{|∇V(x)^T|} \cdot \dot{x} \cdot Φ(∇V(x)^T \dot{x}))\frac{∇V(x)}{|∇V(x)|}] = (n^t \dot{x})$

3- assuming that an upper bound can be placed on the speed:
$$|\dot{x}| ≤ v_{max} \quad (27)$$
the norm of the matrix C may be bound as:
$$|C(x,\dot{x})| ≤ c_{max} \quad (28)$$
4- any inertia matrix belonging to a physical system is positive definite, invertible, and have a bounded norm:
$$|D(x)| ≤ d_{max} \quad (29)$$

where $d_{max}$, $c_{max}$, and $v_{max}$ are finite, positive constants. Based on the above, a dynamic equation that yields an upper bound on δ is:
$$d_{max} \cdot n^T\ddot{x} - c_{max} n^T\dot{x} + B_d \cdot n^T\dot{x} = 0 \quad (30)$$
or
$$\ddot{δ} + Δ \cdot \dot{δ} = 0$$
where $\ddot{δ} = n^T\ddot{x}$, $\dot{δ} = n^T\dot{x}$, and $Δ = \frac{B_d - c_{max}}{d_{max}}$.

To determine the effect of the disruptive time component (ξ(t)) that acts normal to ∇V, the impulse response (h(t)) of (30) is obtained:
$$h(t) = \frac{1}{Δ}(1 - e^{-Δt})Φ(t) = \frac{h'(t)}{Δ} \quad (31)$$
The deviation as a function of time may be computed as:
$$δ(t) = ξ(t) * h(t)$$
where * denotes the convolution operation. Since it was shown in proposition-1 that motion will converge to $x_T$ and all dynamic terms will tend to zero, ξ(t) may be bounded as:
$$\int_0^∞ |ξ(t)|dt ≤ I \quad , \quad (32)$$
therefore: $δ(t) = \frac{1}{Δ} h'(t) * ξ(t) ≤ \frac{I_{max}}{Δ}$,

where I and $I_{max}$ are positive constants. By properly selecting a value for Δ, the maximum deviation $δ_m$ can be made arbitrarily small. In other words the dynamic trajectory of (11) will closely follow the kinematic trajectory of (4) and the spatial constraints will be preserved. It ought to be mentioned that since NADF is by design made to be zero when motion is in accordance with the guidance field ∇V, $B_d$ can be made arbitrarily large without slowing down the system. This fact is clearly reflected by the simulation results (figure-24).

V. Systems with External Forces

The NADF approach may be adapted for designing constrained motion controller for mechanical systems experiencing external forces (e.g. gravity). The dynamical equation of such systems has the form:
$$D(x)\ddot{x} + C(x,\dot{x})\dot{x} + G(x) = F \quad (33)$$
where G(x) and F are vectors containing the external forces and the applied control forces respectively. A controller consisting of the gradient guidance field and a strong enough NADF (34) has the ability to make the trajectory of the system in (33) closely follow the kinematic trajectory from an initial starting point ($x_o$) to the target point $x_T$,
$$F = -B_d \cdot M(x,\dot{x}) - K \cdot ∇V(x) \quad (34)$$
However, due to the presence of the external forces the controller will not be able to hold the state close to the target point and drift will occur (Figure-26). Arimoto and Miyazaki showed that steady state error caused by the external forces may be cancelled by using an integral control action [27]. Unfortunately, an integral action raises the order of the mechanical system and could cause it to become unstable if it is not tuned properly. The integrator also induces difficult to manage transients in the response of the system.

Here an alternative approach to using an integrator is suggested. The suggested approach does not endanger stability and can cancel the error caused by the external forces bringing the dynamic trajectory arbitrarily close to the target point. The approach capitulates on the ability of the controller in (34) to drive motion arbitrarily close to the target point. Once the trajectory is close to the target, a passive clamping control action is activated to trap the trajectory in a set close to the target. After motion is trapped by the clamping control, an iterative procedure is suggested for totally cancelling the error. In the following the suggested clamping control is described

1. Clamping control:

The effect of the clamping control ($F_c$) is strictly localized to a hyper space of radius σ surrounding the target point. If motion is heading towards the target, this control component is inactive. On the other hand, if motion starts heading away from the target, the control becomes active and attempts to drive the trajectory back to the target (Figure-13).

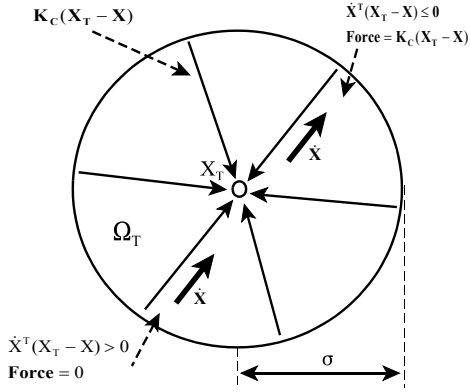

Figure-13: The clamping control.

A form of a clamping control that behaves in the above manner is:

$$F_C(x, \dot{x}) = (x - x_T) \cdot \Phi(\sigma - |x - x_T|) \cdot \Phi(\dot{x}^T(x - x_T)) \quad (35)$$

The strength of $F_c$ is adjusted by multiplying it with a constant $K_c$ so that the steady state error is kept below a desired level ($\epsilon$). Unlike the integrator, the use of a clamping control will keep the mechanical system stable for any positive value of $K_c$.

Proposition-3:
For the mechanical system in (33), a controller of the form:
$$F = -B_d \cdot M(x, \dot{x}) - K \cdot \nabla V(x) - K_C \cdot F_C(\dot{x}, x) \quad (36)$$
can make
$$\lim_{t \to \infty} |x(t) - x_T| \leq \varepsilon < \sigma$$
and
$$\lim_{t \to \infty} \dot{x} = 0 \quad (37)$$

provided that:
1- K, $B_d$, and $K_c$ are all positive,

2- $\quad K_c \geq F_{max}/\epsilon,$
$\quad F_{max} = \max_X |G(x)| \quad x \in \Omega_\sigma$
and $\quad \Omega_\sigma = \{x: |x - x_T| \leq \sigma\} \quad . \quad (38)$

3- a high enough value of $B_d$ is selected so that at some instant in time t`
$$|x(t`) - x_T| < \sigma \quad (39)$$

4- K is high enough so that the gradient field is capable of directing the trajectory to $\Omega_\sigma$

$$\left| K \cdot \nabla V(X) \right| > \left| G^T(X) \frac{\nabla V(X)}{|\nabla V(X)|} \right| \quad X \in \Omega - \Omega_\sigma \quad (40)$$

Proof: Consider a Liapunov function candidate similar to the one in (13) with a gravitational potential energy term (P(X)) added:

$$\Xi(x, \dot{x}) = K \cdot V(x) + \frac{1}{2} \dot{x}^T D(x) \dot{x} + P(x) \quad (41)$$

note that: $\quad G(x) = -\nabla P(x) \quad$ and $\quad P(x) = \int_{x_0}^{x} G(z) \cdot dz \quad . \quad (42)$

Differentiating (42) with respect to time we get:
$$\dot{\Xi}(x, \dot{x}) = K \cdot \nabla V(x)^T \dot{x} + \frac{1}{2} \dot{x}^T \dot{D}(x) \dot{x} + \dot{x}^T D(x) \ddot{x} + \dot{x}^T G(x) \quad (43)$$

solving for $\ddot{x}$ from equations (33, 34) and substituting the results in (43) we get:
$$\begin{aligned}\dot{\Xi} = &- B_d \cdot (n^T \dot{x})^T (n^T \dot{x}) \\ &- B_d \cdot \frac{(\nabla V(x)^T \cdot \dot{x})^T}{|\nabla V(x)|} \cdot \frac{(\nabla V(x)^T \cdot \dot{x})}{|\nabla V(x)|} \cdot \Phi(\nabla V(x)^T \dot{x}) \\ &- K_C \cdot \dot{x}^T (x - x_T) \cdot \Phi(\dot{x}^T (x - x_T)) \cdot F(\sigma - |x - x_T|)\end{aligned} \quad (44)$$

Since $K_c$ and $B_d$ are positive we have:
$$\dot{\Xi} \leq 0 \quad \forall \quad x, \dot{x}, \quad (45)$$
where $\quad \dot{\Xi} = 0 \quad$ for $\quad x, \dot{x} = 0$.

Since we are assuming that K and $B_d$ are selected high enough so that the dynamic trajectory will follow the kinematic trajectory and enter $\Omega_\sigma$, the minimum invariant set to which the trajectory is going to converge may be computed from the equation:
$$G(x) + K \cdot \nabla V(x) + K_C \cdot F_C(x, \dot{x} = 0) = 0 \quad (46)$$

Since $\Phi(0)=1$, and $x \in \Omega_\sigma$ (i.e. $\Phi(\sigma-|x-x_T|)=1$), equation (46) becomes:
$$G(x) + K \cdot \nabla V(x) + K_C \cdot (x - x_T) = 0 \quad (47)$$

As can be seen if condition 2 on $K_C$ is satisfied, the solution of the above equation has to lie in the set $\Omega_\epsilon = \{x: |x-x_T| < \epsilon\}$. This means that the deviation of the end of the dynamic trajectory from the target point should at most be $\epsilon$.

Another alternative to the use of integration is to reduce steady state by increasing the gain of the gradient field (K) to a sufficiently high level. This approach makes the transient difficult to manage and increases the control effort. On the other hand, selecting a high gain of the clamping control ($K_C$) to manage the steady state error will not cause the above problems. This is due to the fact that this control component is designed to be minimally intrusive affecting the system only when it is needed. This is clearly demonstrated by simulation (Figure-27,28,29)

2. Iterative, blind error cancellation:

While clamping control has the ability to reduce the steady state error to an arbitrarily small value, sometimes it is desired that this error be totally cancelled. Here, an iterative, blind procedure is suggested for error cancellation. The procedure works by providing an alternative path (β) other than the error channel ($K_P \cdot e$, where $K_P$ is a positive definite matrix) to supply the control signal (u) that is needed to hold the robot at a location $x_T$ (figure-14),

$$u = k \cdot e + \beta \quad (48)$$

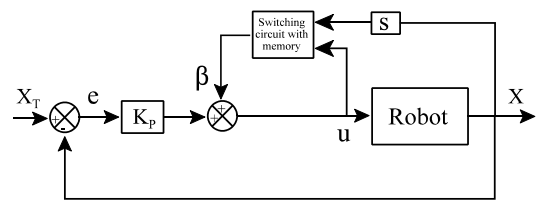

Figure-14: The suggested scheme for iterative error cancellation.

The fixed point iteration method [28] is used to evolve an estimate of the control signal so that the steady state error is driven to zero. This procedure is implemented using a switched logic circuit with one memory storage element. One implementation requires the circuit to have two inputs: the control that is directly fed to the robot and velocity of the robot's coordinates in order to assess convergence (other means to decide if the robot has converged may be used). There is only one output consisting of the bias term $\beta$. The bias term is iterativly determined as follows: when motion is about to settle (i.e. $|dx/dt| < \alpha$, where $0 < \alpha << 1$), the circuit measures the value of u and assigns it to v. This value is kept till at another instant i the event becomes true again. At the i'th instant we have:

$$u = G(x_i), \quad \beta = G(x_{i-1}), \text{ and } K_P \cdot e = K_P \cdot (x_T - x_i) \quad (49)$$

where $x_i$ is the position of the robot at the i'th settling instant. Relating the above quantities using (48) yields the recursive relation:

$$G(x_i) = G(x_{i-1}) + K_P \cdot (x_T - x_i). \quad (50)$$

Proposition-4:
The recursive relation in (50) has a fixed point at which:

$$(x_T - x_i) = 0 \quad (51)$$

Proof: Using Taylor series expansion around $x_T$, we have:

$$G(x) = G(x_T) + J(G(x_T))(x - x_T) + .... \quad (52)$$
$$= G(x_T) + F(x - x_T)$$

where J is the Jacobian matrix of G and F is a function containing the $(x-x_T)$ terms of the Taylor series. Substituting (52) into (50) we get:

$$F(e`_i) = F(e`_{i-1}) - K_P \cdot e`_i \quad (53)$$
where $\quad e`_i = -(x_T - x_i). \quad (54)$

Now let $\eta = F(e`)$ and Q be the inverse function of F in the neighborhood of $x_T$. Substituting Q in (53), we obtain the recursive relation:

$$K_P \cdot Q(\eta_i) + \eta_i = \eta_{i-1}. \quad (55)$$

At a fixed point we have:

$$\eta_i = \eta_{i-1} \quad (56)$$
or $\quad K_P \cdot Q(\eta_i) = 0.$

Since $K_P$ is positive definite, i.e. it is not singular:
$$Q(\eta_i) = e`_i = (x_i - x_T) = 0 \quad (57)$$
In other words: $\quad x_i = x_T$.

Proposition-5:
For any positive definite $K_P$, the fixed point $x = x_T$ is a stable attractor fixed point, i.e. if $x_i$ is sufficiently close to $x_T$,
$$\lim_{i \to \infty} \mathbf{x_i} \to \mathbf{x_T} \quad (58)$$
Proof: In the close neighborhood of $x_T$, equation (50) may be written as:

$$J(G(x_T)) \cdot (x_i - x_T) = J(G(x_T)) \cdot (x_{i-1} - x_T) + K_P \cdot (x_T - x_i) \quad (59)$$
Notice that: $\quad J(G(x_T)) = J(\nabla P(x_T)) = H(x_T) \quad (60)$

where H is the symmetric hessian matrix. Substituting (60) in (59) yields the equation:

$$[K_P + H(x_T)] \cdot e_i = H(x_T) \cdot e_{i-1} \quad (61)$$
where $\quad e_i = (x_T - x_i).$

Since $K_P$ is positive definite and H is symmetric, they are simultaneously diagonalizable into:

$$K = U U^T \text{ and } H = U \Lambda U^T \quad (62)$$

where U is a nonsingular matrix and $\Lambda$ is a diagonal matrix with non-negative elements $\lambda_l$, $l=1,..,N$, see [29, page-86]. Using the above decomposition (61) may be written as:

$$U(I + \Lambda) U^T \cdot e_i = U \Lambda U^T \cdot e_{i-1} \quad (63)$$

Using the transformation $q_i = U^T \cdot e_i$,
we have $\quad q_i = A \cdot q_{i-1} \quad (64)$

where

$$\mathbf{A} = (\mathbf{I} + \Lambda)^{-1} \Lambda = \begin{bmatrix} \frac{\lambda_1}{1+\lambda_1} & 0 & \cdot & 0 \\ 0 & \frac{\lambda_2}{1+\lambda_2} & \cdot & 0 \\ \cdot & \cdot & \cdot & \cdot \\ 0 & 0 & \cdot & \frac{\lambda_N}{1+\lambda_N} \end{bmatrix}. \quad (65)$$

It is well-known that the solution of (64) is:
$$q_i = A^i \cdot q_0 \quad (66)$$
Since
$$0 \leq \frac{\lambda_l}{1+\lambda_l} < 1 \quad l=1,..,N \quad (67)$$

we have: $\quad \lim_{i \to \infty} \mathbf{q_i} = \lim_{i \to \infty} \mathbf{U^T \cdot e_i} \to \mathbf{0}. \quad (68)$

Since U is a nonsingular matrix

$$\lim_{i \to \infty} \mathbf{e_i} \to \mathbf{0} \quad (69)$$
or $\quad \lim_{i \to \infty} \mathbf{x_i} \to \mathbf{x_T} \quad (70)$

VI. Results

In this section several simulation examples are presented to demonstrate the capabilities and versatility of the NADF approach.

**1. Point mass in a cluttered environment:**
The gradient field in figure-2 is augmented with NADF instead of the linear, viscous, dampening forces. The combination of both gradient field and NADF is used to steer a 1Kg mass from a start point to a target point. An excessively high dampening coefficient, $B_d=10$, is used. The trajectory of the mass is shown in figure-15, and the mass distance to the target, D(t), as a function of time is shown in figure-16. As can be seen, the kinodynamic trajectory of the mass is almost identical to that marked by the gradient field (kinematics only) in figure-3. Moreover, motion of the mass is almost six times faster than its viscous dampening counterpart shown in figure-7 with a settling time ($T_S$) of about 12 seconds compared to 72 seconds. Figure-17 shows the corresponding radial speed along the mass trajectory and figure-18 shows the control signal (X-Y force components).

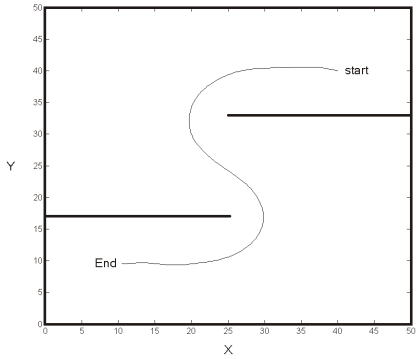

Figure-15: Trajectory, NADF, Bd=10.

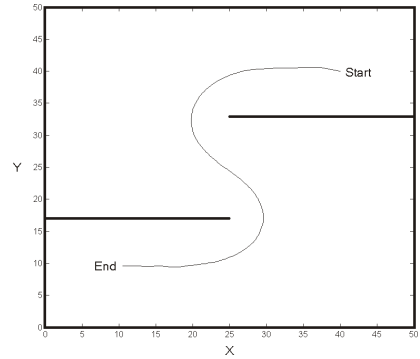

Figure-19: Trajectory - speed limit imposed.

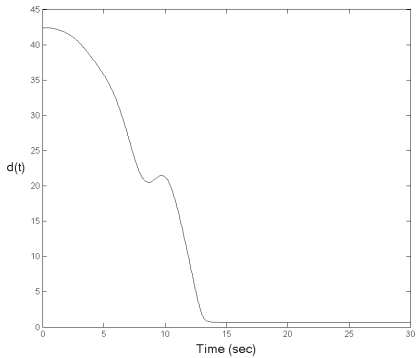

Figure-16: Distance to target versus time, NADF.

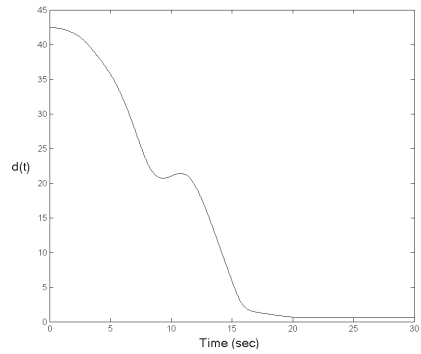

Figure-20: Distance to target versus time - speed limit imposed.

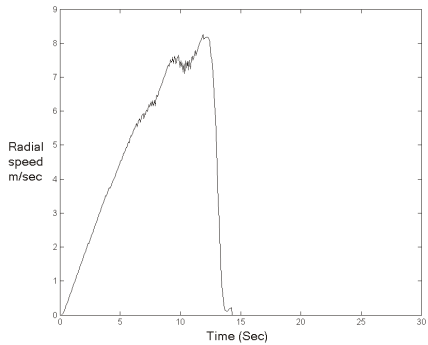

Figure-17: Radial speed versus time, NADF.

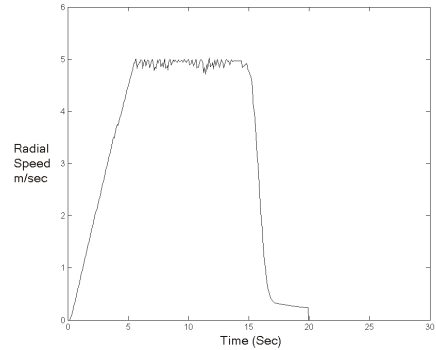

Figure-21: Radial speed of the trajectory - speed limits imposed.

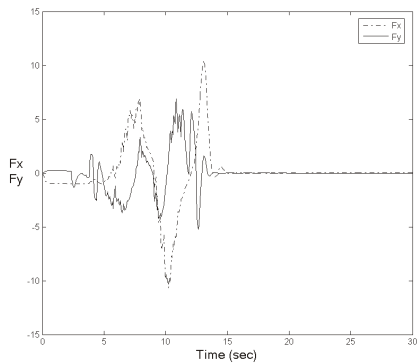

Figure-18: x and y control force components, NADF.

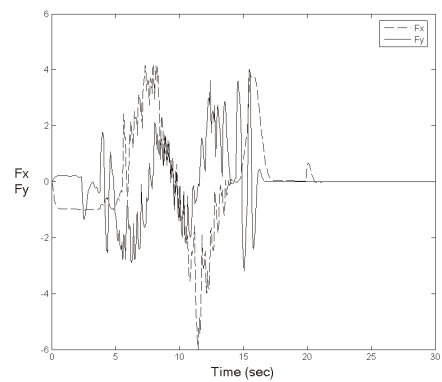

Figure-22: x and y control force components for the point mass with constraints on speed.

## 2. Point mass with upper speed limit:

As can be seen from figure-17, the radial speed of the mass continuously increases prior to reaching the target. This is a result of NADF not having any friction forces along the radial component of the trajectory. While a fast response is required, measurers to prevent motion from racing out of control are desirable. Placing an upper limit on speed may be done using the approach in [4]. In figure-19 a 5 m/sec speed limit is imposed on the trajectory. As can be seen from figure-21, the radial speed keeps increasing and stops at the imposed limit. While limiting speed causes the settling time to increase from 12 seconds to 16 seconds (figure-20), it also causes a drop in the control effort (figure-22).

## 3. Settling time - a comparison:

NADF and linear viscous dampening exhibit fundamentally different behavior as far as convergence is considered. The settling time for the point mass with no constraints on speed example is drawn in figure-23 as a function of the linear viscous friction coefficient (B). As can be seen, the $T_S$-B relation is convex with one value for B corresponding to a global minimum of $T_S$. This is expected since for low B high oscillations will prevent motion from quickly settling in the 5% zone around the target. On the other hand, a high value for B reduces the oscillations by slowing down the response delaying the entrance to the 5% zone.

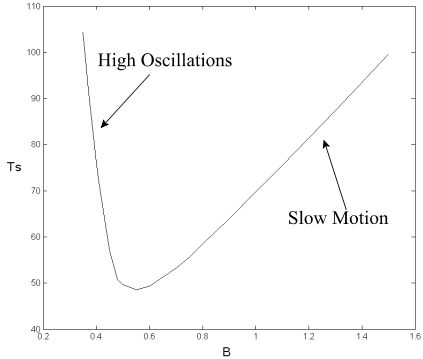

Figure-23: $T_S$ versus B for linear dampening.

The relation between $T_S$ and the coefficient of NADF ($B_d$) is a rapidly and strictly decreasing one (figure-24). Similar to the linear case, for a low value of $B_d$ high oscillations will prevent the quick capture of the trajectory in the 5% zone around the target. As the value of $B_d$ increases, NADF, by design, only impedes the component of motion along the coordinate field tangent to the gradient guidance field (figure-25). This component does not contribute to convergence and it only causes delay in reaching the target. Since NADF attenuates this and only this component of motion leaving the motion along the gradient field unaffected, the delay in reaching the target drops as $B_d$ increases yielding a strictly decreasing profile of the $T_S$-$B_d$ curve.

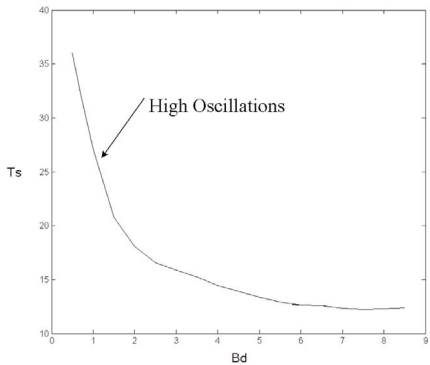

Figure-24: Settling time versus NADF coefficient.

The $T_S$ versus the coefficient of dampening profile is important. It determines the ability to tune the controller so that the specifications are met. In tuning the controller there are two requirements: it is required that the maximum spatial deviation ($\delta_m$) between the kinematic and the dynamic path be as small as possible so that the constraints are upheld. It is also required that the settling time be as small as possible. The first requirement is achieved by making the coefficient of dampening high enough. In the linear viscous dampening case one can only strike a compromise between $T_S$ and $\delta_m$. For the NADF case this compromise is not needed since both $T_S$ and $\delta_m$ are strictly decreasing as a function of $B_d$.

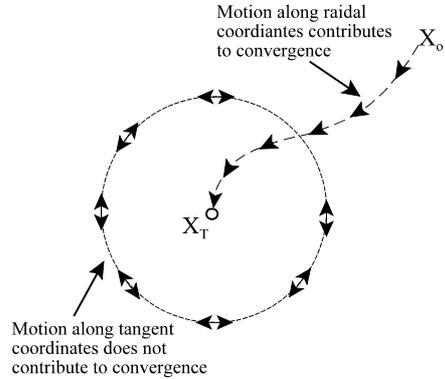

Figure-25: Motion along gradient lines contribute to convergence, motion along tangent liens causes only delays.

## 4. Point mass with external forces

As mentioned before, the NADF approach may be adapted to work with second order systems experiencing external forces using the suggested clamping control. In this example a point mass with constant external forces acting on it having the system equation in (71) is controlled using a gradient field and NADF.

$$\begin{bmatrix} \ddot{\mathbf{x}} \\ \ddot{\mathbf{y}} \end{bmatrix} = \begin{bmatrix} -4 \\ -4 \end{bmatrix} \quad (71)$$

As can be seen from figure-26, for a sufficiently high $B_d$ the controller will succeed in driving the mass to the target and avoiding the obstacles. However, when the target is reached, drift caused by the external forces occur.

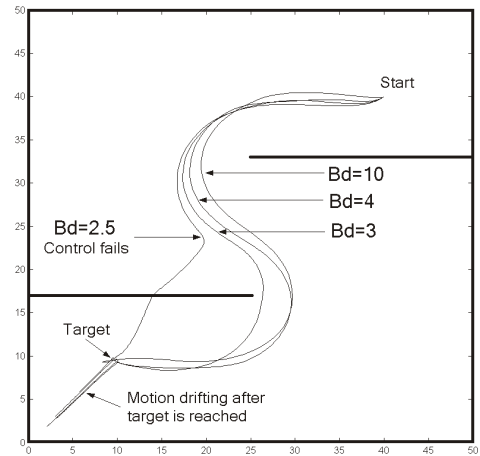

Figure-26: Trajectory, NADF - external force present.

In figure-27 a clamping control similar to the one in (35) is added with K=1, $B_d$=10, $K_C$=10. As can be seen, the controller was able to hold the trajectory near the target point relying only on a loose, upper bound estimate of the drift. Despite the high value of $K_C$, the trajectory settled in an overdamped manner with no oscillations taking place. The distance versus time curve and the Fx, Fy control forces are shown in figures-28, 29 respectively.

The sliding mode (SM) control approach suggested by Guldner and Utkin in [21] for converting a gradient guidance signal in to a control signal has the ability to handle systems with external forces. In this approach a sliding surface (S) is defined as:

$$S = \dot{x} - v_d(t)\frac{-\nabla V}{|-\nabla V|} \quad . \quad (72)$$

Using this surface a control signal is constructed as:

$$F = -F_0 \frac{S}{|S|} \quad (73)$$

where $v_d$ and $F_o$ are the maximum allowable speed and control forces respectively. The sliding mode control is applied to the point mass with drift in (71). The parameters of the sliding surface are set so that a settling time of 6 sec similar to the one using NADF and clamping control is obtained. $F_o$ is set to obtain a maximum control effort of 100 N. The trajectory is shown in figure-30, the distance to the target versus time is in figure-31. The control forces are shown in figures-32,33. Compared to the NADF approach with clamping the trajectory obtained using the SM approach is a little shaky and experiences some oscillations near the target. However, the biggest difference has to do with the quality and magnitude of the control signals used by both approaches.

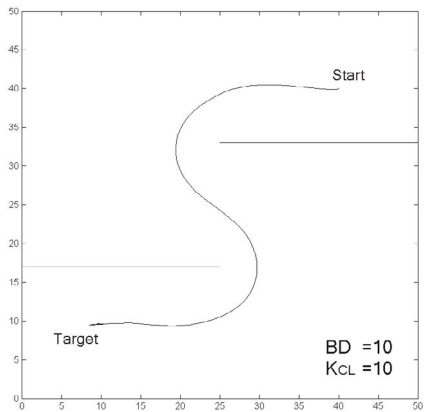

Figure-27: Trajectory, NADF and clamping - external force present.

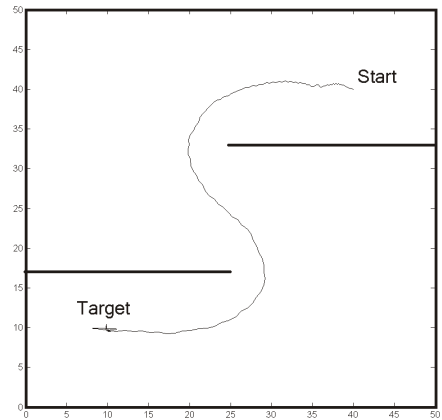

Figure-30: Trajectory - sliding mode control.

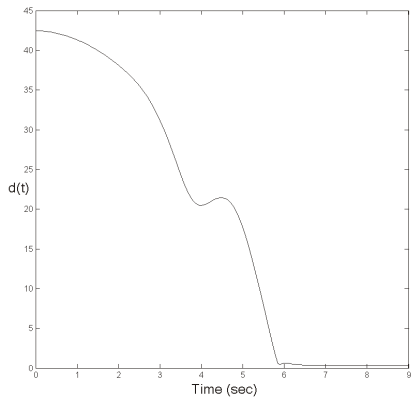

Figure-28: Distance to target versus time - NADF and clamping - external force present.

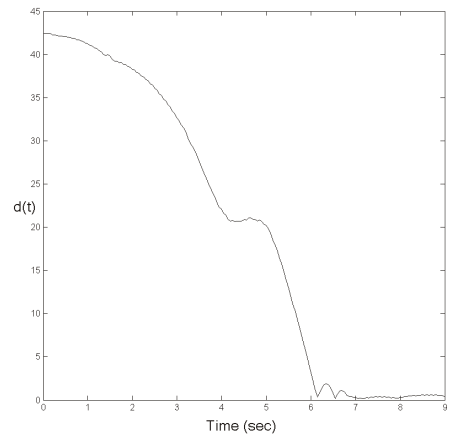

Figure-31: Distnace to target versus time - sliding mode control.

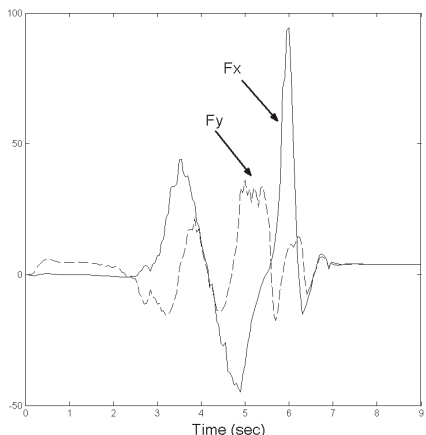

Figure-29: x and y control force components, NADF and clamping - external force present.

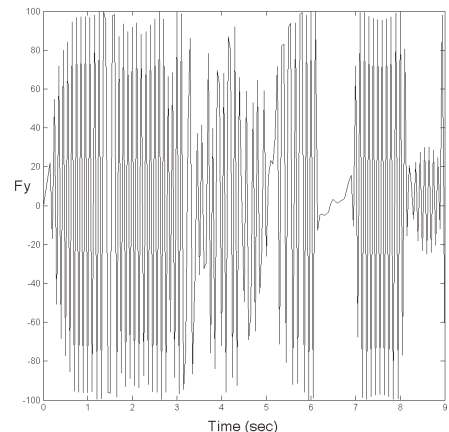

Figure-32- y force control component - sliding mode control.

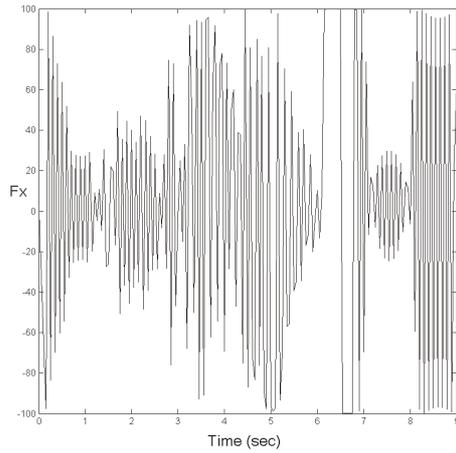

Figure-33: x force control component - sliding mode control.

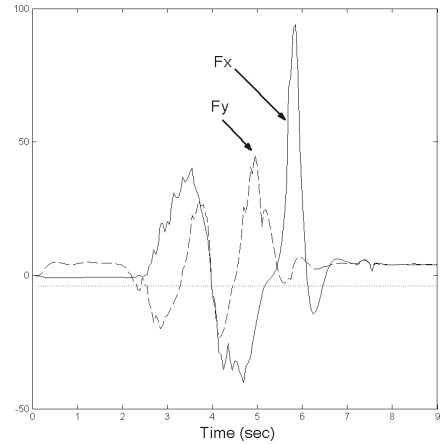

Figure-36: x and y control force component with sensor noise present.

**5. Point mass with external forces and sensor noise:**
The NADF is robust to the presence of sensor noise. The example in figure-34 of a point mass with drift is repeated with sensor noise causing errors in localizing the boundary of the environment. The trajectory, distance to the target versus time curve, and control forces are shown in figures-35,36. Despite the considerable amount of noise and the fact that the raw sensory data was used in the synthesis of the controller without any processing, the trajectory, convergence characteristics and control signal remained reasonably well-behaved.

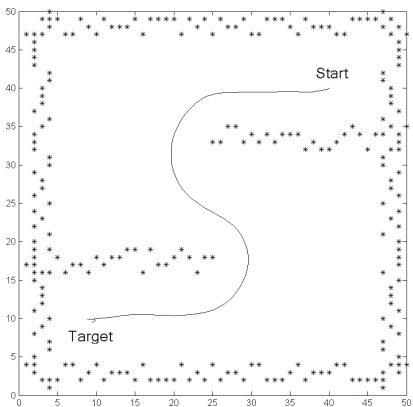

Figure-34: Sensor noise did not have too much effect on the trajectory.

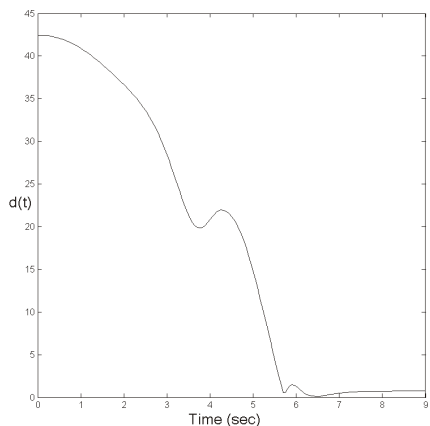

Figure-35: Distance to target versus time (sensory noise).

**6. The narrow corridor effect:**
In his seminal work that appeared in the mid-eighties [9] Khatib suggested that the sensors of a robot be directly coupled to its servo loops. The coupling was achieved via potential fields. The result was a very high increase in the speed at which the robot responds to the contents of the environment.

In the early nineties, Koren and Bornestien reported what they referred to as a serious and inherent deficiency in Khatib's method [15]. They found that if an autonomous robot that is guided by the potential field method is operating in a narrow corridor, the robot could behave erratically, oscillating in a sustained manner between the two walls of the corridor. The artifact was called the "narrow corridor effect". The implications of such a finding are significant. Since a service autonomous robot will have to pass through corridors in order to deliver mail in offices, laundry in hospitals, or parts in factories, the use of potential field-based planners is immediately ruled out and alternatives should be sought.

While this author agrees with [15] that the narrow corridor effect is a serious artifact, he disagrees with it being an inherent deficiency in the potential field approach. This artifact is caused by a misunderstanding of the dual role the gradient of a potential field plays as both a control and guidance provider. This misunderstanding led to an improper coupling between the gradient field and the robot's servo loops that, among other things, caused the narrow corridor artifact. The NADF approach suggested in this paper takes this fact into account and properly couples the gradient field to the servo loops of a robot totally eliminating the narrow corridor effect.

A mobile robot utilizing Khatib's approach behaves normally in an empty corridor (figure-37). However, its behavior changes dramatically if an obstruction is present along its way. The presence of the obstruction seems to ignite sustained oscillations in the trajectory of the robot (figure-38).

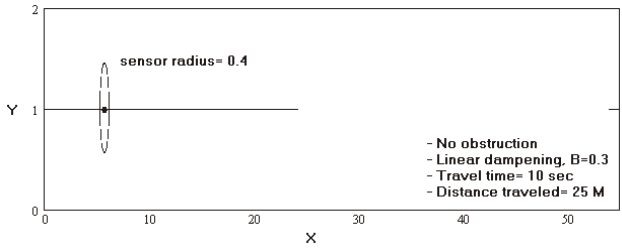

Figure-37: Motion proceeds normally when the corridor is empty.

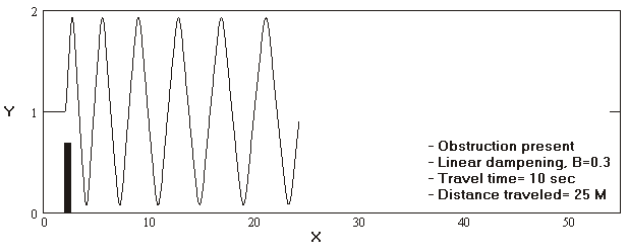

Figure-38: The presence of an obstruction triggers sustained oscillations.

In figure-39 the coefficient of dampening is increased ten times (from B=.3 to B=3) in order to get rid of the oscillations. As can be seen, remaining strong transients are clear. In the previous case, the robot was able to travel 25 meters in 10 seconds. The increase in dampening cut the travel distance to 4 meters in 10 seconds making the robot impractically slow.

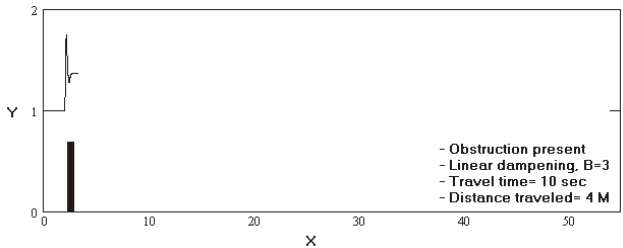

Figure-39: Increasing linear dampening slowed down the system and did not eliminate transient.

The linear viscous dampening force is replaced by NADF. The dampening coefficient used is $B_d$=5 (figure-40). As can be seen, the robot responded well to the presence of the obstruction with little overshoot taking place. Not only a significant improvement in transients was achieved, the robot, despite the large value of the dampening coefficient, became more agile covering more than twice the distance in the linear dampening case (figure-37). A significant increase of $B_d$ to 30 seems to have no effect on the distance the robot is able to travel (figure-41).

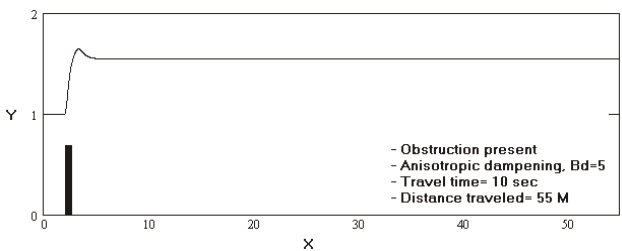

Figure-40: NADF significantly reduced transients and speeded up motion.

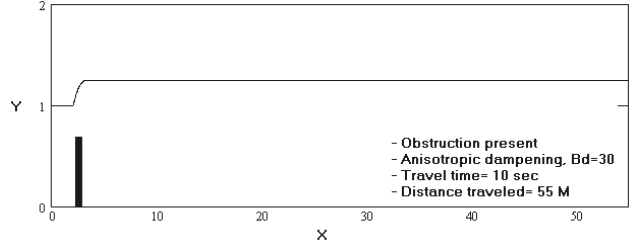

Figure-41: Increasing Bd did not slow down motion.

In figure-42 robustness of the approach to the presence of sensor noise is tested. A wideband noise that is uniformly distributed between (-0.5, 0.5) is added to the sensor causing uniform jitters in the registered reading of the wall. Same as figure-41, a Bd=30 is used. As can be seen, the effect of this relatively large sensor noise is almost negligible on the trajectory of the robot where a steady path was still maintained and the travel distance was not affected.

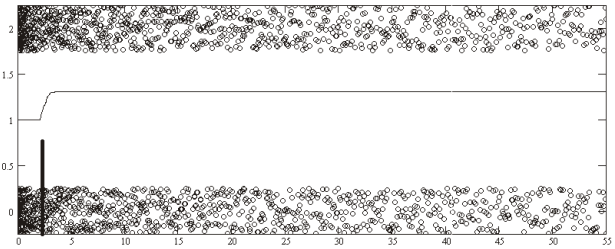

Figure-42: Same as figure-40 but with sensor noise added.

In figure-43 the ability of the NADF control to handle emergency braking of motion is tested. A barrier is placed in the path of the robot. As can be seen, the control was able to brake motion in a well-behaved manner. Figure-44 shows the control forces. Figure-45 demonstrates the ability of the controller to handle multiple obstructions.

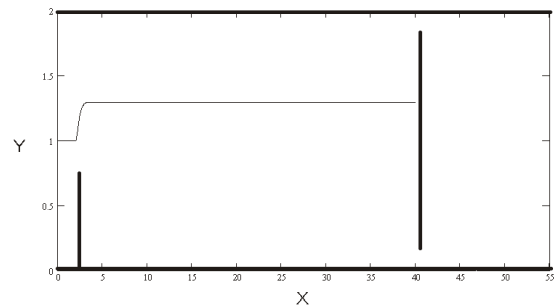

Figure-43: Robot braking motion to avoid collision.

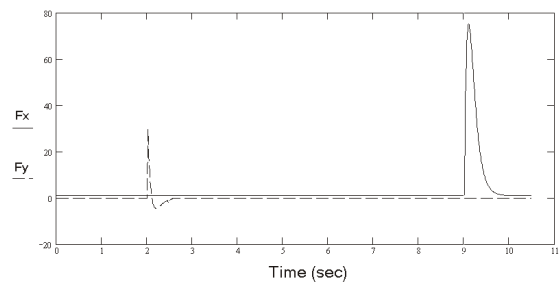

Figure-44: x and y components of the braking force.

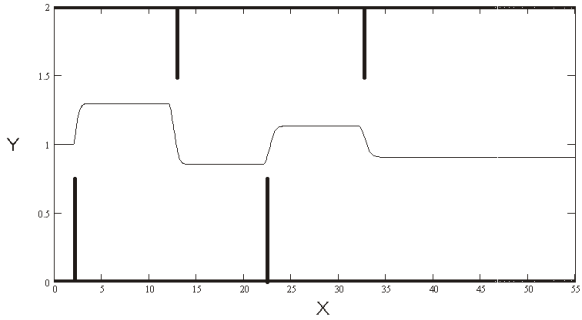

Figure-45: he robot passing through a multi-obstruction environment.

## 7. The multi-agent case:

NADFs may also be applied for the multi-robot case. In [22,30] a complete, decentralized, multi-agent planner was suggested considering the kinematics of the robots only (figure-46).

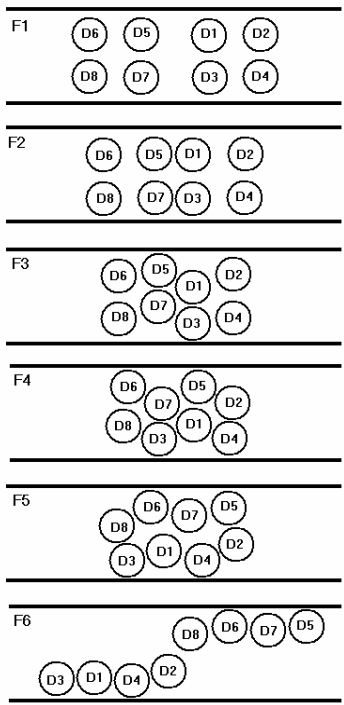

Figure-46: Self-organization using the vector-harmonic potential approach.

Figure-47 shows the paths for two massless robots that are trying to exchange positions. When mass is added (1Kg each), the planner totally fails (figure-48).

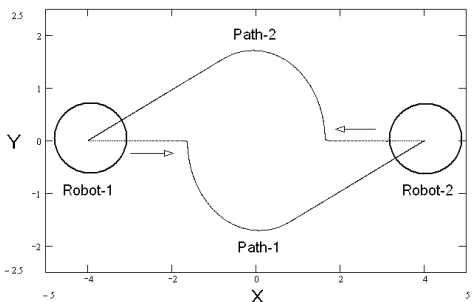

Figure-47: Two robots exchanging positions, kinematics only.

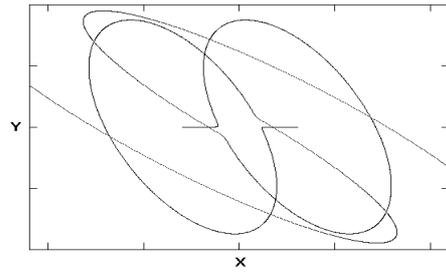

Figure-48: Same as figure-14, but with 1kg mass added to each robot.

In figure-49 linear dampening is added to control the inertial forces (B=1). Figure-50 shows the distance to target of robot-1 as a function of time. It took the robot about 13 seconds to reach its target. In figure-51, NADF was used ($B_d$=10). As can be seen from the distance - time profile in figure-52, it took robot-1 only two seconds to reach its target.

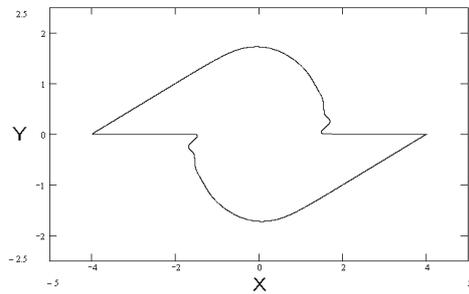

Figure-49: Linear dampening added to manage inertia, B=1.

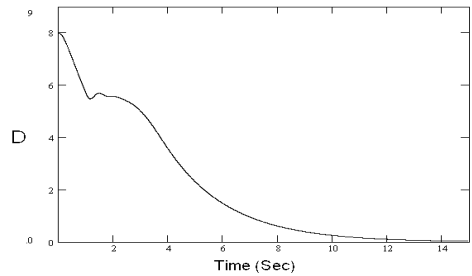

Figure-50: Distance to target versus time for robot-1 in figure-16.

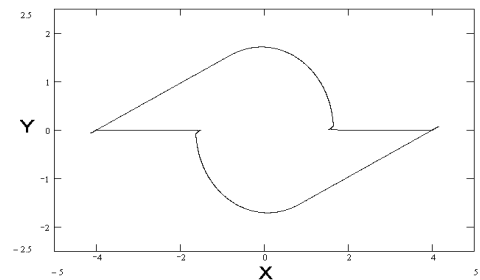

Figure-51: NADF added to manage inertia, Bd=10.

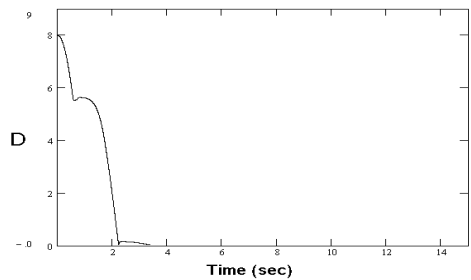

Figure-52: Distance to target versus time for robot-1 in figure-52.

## 8. Iterative error removal:

The iterative procedure to remove the steady state error suggested in the previous section is tested using a simple pendulum (figure-53) with concentrated mass M=1Kg and length L=1M. The dynamic equation of the pendulum is:

$$\mathbf{M} \cdot \mathbf{L} \cdot \ddot{\Theta} + \mathbf{M} \cdot \mathbf{g} \cdot \sin(\Theta) = \mathbf{u} \qquad (74)$$

where g is the acceleration constant and u is the external applied control torque.

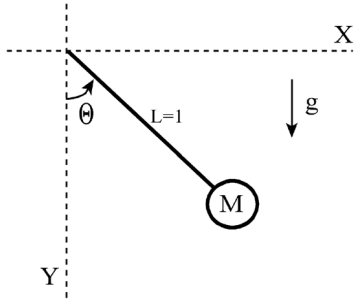

Figure-53: A simple pendulum.

A simple controller with position and velocity feedback (75) is used to move the pendulum from $\Theta=0$ to $\Theta=\pi/2$.

$$\mathbf{u} = -\mathbf{K} \cdot \Theta - \mathbf{B} \cdot \dot{\Theta} \qquad (75)$$

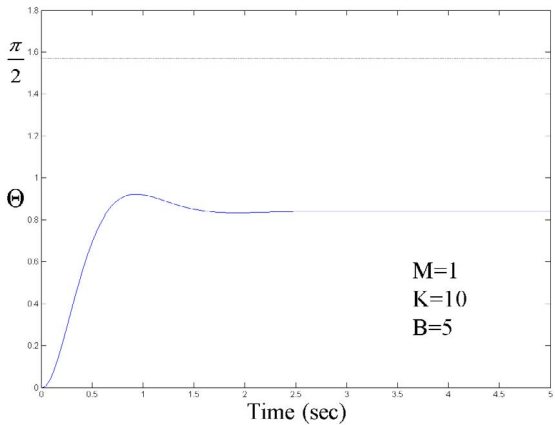

Figure-54: Steady state error caused by weight of pendulum.

As can be seen from figure-54, the weight of the pendulum causes significant steady state error. In order to remove the error, the switching circuit suggested in V.2 is added to the controller. Different switching thresholds are used to assess the sensitivity of the procedure to the presence of transients (figure-55). As can be seen, the error was eliminated in all cases. Although the iterative error cancellation procedure was designed to be used when transients fade away and motion settles, simulation shows that the procedure exhibits little sensitivity to the presence of transients that enables us to loosely choose the threshold α. Actually, the simulation reveals that better results in terms of having a lower settling time could be obtained if switching is carried out before motion completely settles.

In figure-56 the effect of the forward gain on the speed of convergence is shown. As expected, the higher the forward gain is the faster the system converges to its target.

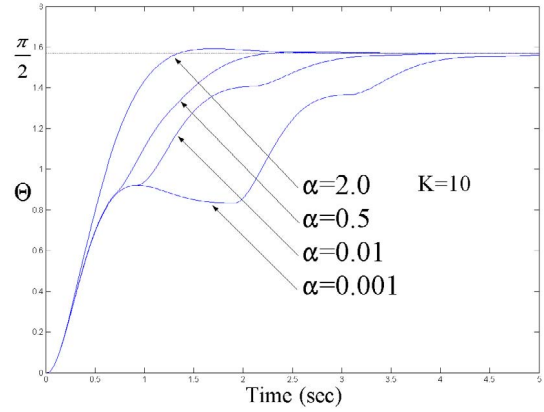

Figure-55: Error cancellation using switching circuit - different thresholds.

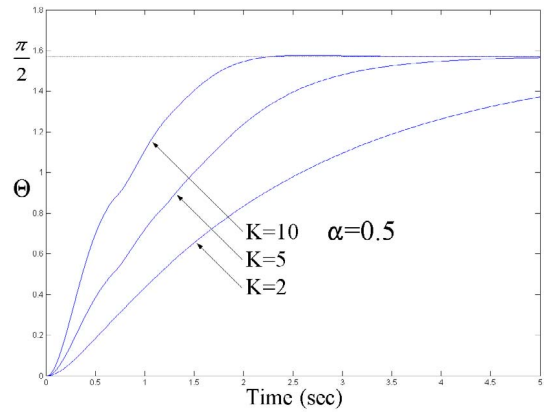

Figure-56: Effect of forward gain on convergence.

## VII. Conclusions

In this paper the capabilities of the HPF approach are extended to tackle the kinodynamic planning case. The extension is provably-correct and bypasses many of the problems encountered by previous approaches. It is based on a novel type of nonlinear, passive dampening forces called NADFs. The suggested approach enjoys several attractive properties. It is easy to tune; it can generate a well-behaved control signal; the approach is flexible and may be applied in a variety of situations, it is provably-correct; it is resistant to sensor noise; it does not require exact knowledge of system dynamics, and it can tackle dissipative systems as well as systems under the influence of external forces.

It ought to be emphasized that most of the problems attributed to the potential field approach are a result of the misunderstanding of the dual role a potential field plays as a motion actuator and a guidance provider. The NADF approach is a step forward in taking both of these roles into account.

## Acknowledgment

The author would like to thank KFUPM for its support of this work.

## Appendix

A. *Definition*: Let V(X) be a smooth ( at least twice differentiable) scalar function (V(X): $R^N \rightarrow R$). A point Xo is called a critical point of V if the gradient vanishes at that point ($\nabla V(Xo)=0$); otherwise, Xo is regular. A critical point is Morse, if its Hessian matrix (H(Xo)) is nonsingular. V(X) is Morse if all of its critical points are Morse [24].

B. *Proposition*: If V(X) is a harmonic function defined in an N-dimensional space ($R^N$) on an open set Ω, then the Hessian matrix at every critical point of V is nonsingular, i.e. V is Morse.

*Proof*: There are two properties of harmonic functions that are used in the proof:
1- a harmonic function (V(X)) defined on an open set Ω contains no maxima or minima, local or global in Ω. An extrema of V(X) can only occur at the boundary of Ω,

2- if V(X) is constant in any open subset of Ω, then it is constant for all Ω. Other properties of harmonic functions may be found in [26].

Let Xo be a critical point of V(X) inside Ω. Since no maxima or minima of V exist inside Ω, Xo has to be a saddle point. Let V(X) be represented in the neighborhood of Xo using a second order Taylor series expansion:

$$V(X) = V(Xo) + \nabla V(Xo)^T (X - Xo) + \frac{1}{2}(X - Xo)^T H(Xo)(X - Xo) \qquad |X-X0|<<1. \qquad 76$$

Since Xo is a critical point of V, we have:

$$V' = V(X) - V(Xo) = \frac{1}{2}(X - Xo)^T H(Xo)(X - Xo) \qquad |X-X0|<<1. \qquad 77$$

Notice that adding or subtracting a constant from a harmonic function yields another harmonic function, i.e. V` is also harmonic. Using eigenvalue decomposition [25]:

$$V' = \frac{1}{2}(X - Xo)^T U^T \begin{bmatrix} \lambda_1 & 0 & 0 & 0 \\ 0 & \lambda_2 & . & 0 \\ . & . & . & . \\ 0 & 0 & . & \lambda_N \end{bmatrix} U(X - Xo) \qquad 78$$

$$= \frac{1}{2}\xi^T \begin{bmatrix} \lambda_1 & 0 & 0 & 0 \\ 0 & \lambda_2 & . & 0 \\ . & . & . & . \\ 0 & 0 & . & \lambda_N \end{bmatrix} \xi = \frac{1}{2}\sum_{i=1}^{N}\lambda_i \xi_i^2$$

where U is an orthonormal matrix of eigenvectors, $\lambda$'s are the eigenvalues of H(Xo), and $\xi=[\xi_1\ \xi_2\ ..\xi_N]^T = U(X-Xo)$. Since V` is harmonic, it cannot be zero on any open subset Ω; otherwise, it will be zero for all Ω, which is not the case. This can only be true if and only if all the $\lambda_i$'s are nonzero. In other words, the Hessian of V at a critical point Xo is nonsingular. This makes the harmonic function V also a Morse function.

The navigation function defined in [13] is a special case of a harmonic potential field. According to [13] a navigation function must satisfy the following properties:

1- it is smooth (at lest $C^2$),
2- it contains only one minimum located at the target point,
3- it is a Morse function,
4- it is maximal and constant on Γ.

A harmonic function (V) is $C^\infty$ and Morse. Harmonic functions are extrema-free in Ω. Their maxima and minima can only happen at the boundary of Ω. In the harmonic approach Γ and the target point ($X_T$) are treated as the bounary of Ω. Through applying the appropriate boundary conditions, the minimum of V is forced to occur on $X_T$. Also, by the application of the Drichlet boundary conditions, the value of V is forced to be maximal and constant at Γ. The Drichlet condition (constant potential on the boundary) is one of many settings used in constructing a harmonic potential that may be used for navigation.